\documentclass[10pt,twocolumn,letterpaper]{article}
\usepackage[accepted]{cvpr}
\usepackage[dvipsnames]{xcolor}

\usepackage[dvipsnames]{xcolor}

\definecolor{cvprblue}{rgb}{0.21,0.49,0.74}
\usepackage[pagebackref,breaklinks,colorlinks,citecolor=cvprblue]{hyperref}

\usepackage{array}
\newcolumntype{?}{!{\vrule width 1.3pt}}

\def\paperID{44} 
\def\confName{CVPR}
\def\confYear{2024}

\title{Temporally Consistent Object Editing in Videos using Extended Attention}

\author{\parbox{\textwidth}{\centering
AmirHossein Zamani$^{1,2}\;$ \thanks{This research was partially funded by NSERC Discovery Grant RGPIN-2021-04104 and Mila Tech Transfer, correspondence to: {\tt\small amirhossein.zamani@concordia.ca}}\hspace{-8pt}
\qquad
Amir G. Aghdam$^{1}$\hspace{-10pt}
\qquad
Tiberiu Popa$^{1}$\hspace{-10pt}
\qquad
Eugene Belilovsky $^{1,2}$ \hspace{-10pt}
} \vspace{5pt}\\
 $^1$ Concordia University, Montreal, QC, Canada; $^2$ Mila -- Quebec AI Institute \\
}

\begin{document}
\maketitle

\begin{abstract}\vspace{-5pt}
    Image generation and editing have seen a great deal of advancements with the rise of large-scale diffusion models that allow user control of different modalities such as text, mask, depth maps, etc. However, controlled editing of videos still lags behind. 
    Prior work in this area has focused on using 2D diffusion models to globally change the style of an existing video. On the other hand, in many practical applications, editing localized parts of the video is critical. In this work, we propose a method to edit videos using a pre-trained inpainting image diffusion model. We systematically redesign the forward path of the model by replacing the self-attention modules with an extended version of attention modules that creates frame-level dependencies. In this way, we ensure that the edited information will be consistent across all the video frames no matter what the shape and position of the masked area is. We qualitatively compare our results with state-of-the-art in terms of accuracy on several video editing tasks like object retargeting, object replacement, and object removal tasks. Simulations demonstrate the superior performance of the proposed strategy.\vspace{-10pt}
\end{abstract}    
\vspace{-8pt}
\section{Introduction}\vspace{-5pt}
{
   \begin{figure*}[h]
      \centering
      \includegraphics[scale=0.14]{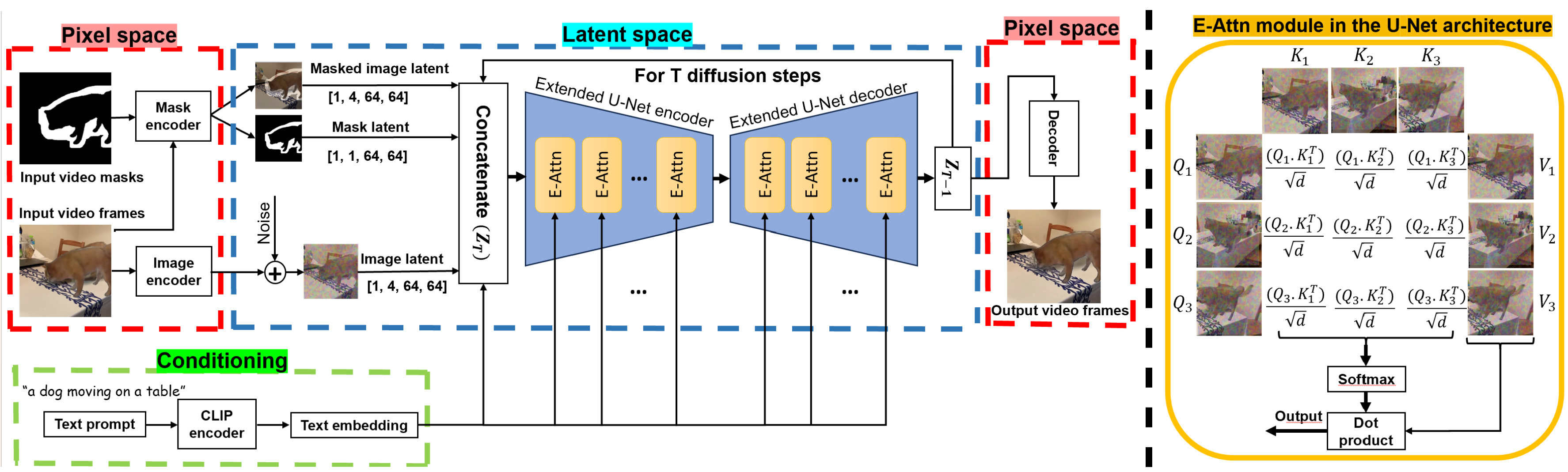}
    
       \caption{An overview of the diffusion process for temporal consistent video editing. To make the video editing process temporally consistent, we extend the U-Net architecture by replacing original attention modules used in \cite{LDMs} with Extended Attention modules \vspace{-10pt}}
       \label{fig:DiffusionProcess}
    \end{figure*}
    
    In recent years, significant strides have been made in large-scale generative modeling with diffusion models, leading to advancements in generating and manipulating image content \cite{DDPM, ControlNet, LDMs}. However, applying these advancements to video editing has been relatively limited, stemming from inconsistencies inherent in the outcomes generated by text-to-image models. Within the domain of video editing and inpainting, a key objective is to ensure visual and temporal coherence in all frames of the edited/generated video. To this end, considering different applications such as video synthesis and video editing, different methods are proposed in the literature that can be categorized into three groups based on what they condition on: 1) Mask-guided methods \cite{ProPainter, E2FGVI, FGT, DMT} in which deep neural models are conditioned on the binary mask of the input video. These methods are also called video inpainting methods in which the goal is to fill the masked (missing) area in the input video. Most of these methods are utilized for objects, watermarks, and logos removal, and video completion tasks.  2) Text-guided methods \cite{TokenFlow, Magicedit, VIDiff}: These approaches usually benefit from a diffusion model \cite{LDMs} conditioned by a text prompt. More specifically, this type of method requires detailed textual descriptions of both the original and target videos and then reconstructs the videos based on these descriptions for video synthesizing and editing purposes. 3) Multi-modal-guided methods: Recent approaches \cite{VideoComposer, ControlVideo, Lumiere} introduce new stronger conditioning for full video generation, such as a style image, pose, depth, and sketch which leads to having more flexibility and control ability for video generation. However, they are not able to support localized editing of videos.   
    
    Despite the progress made in this area, existing methods still suffer from inconsistencies across frames of the generated/synthesized videos. Moreover, all the mask-guided methods assume that the mask by which their model is conditioned has a fixed shape and position in all the frames which is not generally the case in real-world video inpainting applications. Hence, this type of mask-guided method fails to handle mask images whose positions and geometric shapes vary across the frames. Last but not least, among diffusion-based consistent video editing techniques, almost all of them require either fine-tuning or training on large text-image or text-video datasets which practically is costly, time-consuming, and in some situations infeasible. To overcome these issues, inspired by the idea of extended attention layers \cite{TokenFlow, TuneAVideo, Pix2Video}, we design and develop a mask and text video editing framework by leveraging a pre-trained inpainting image diffusion model \cite{LDMs}, allowing novel applications in mask guided video editing. We systematically redesign the forward path of the pre-trained diffusion model by replacing the self-attention modules with an extended version of attention modules without any additional training or fine-tuning. In this way, we ensure that the edited information will be consistent across all the video frames no matter what the shape and position of the masked area is. This approach allows us to use our framework for multiple video editing tasks such as object retargeting, object replacement, and object removal tasks while previous mask-guided methods only focus on the object removal task.
}

\vspace{-5pt}\section{Methodology}\vspace{-5pt}
\label{Methodology}
{   
    Stable Diffusion (SD) \cite{LDMs} is a well-known text-to-image diffusion model that adds/removes the noise to/from the latent (feature) space of the image, not the image itself during the diffusion process. 
    This model can be conditioned on different modalities of signals such as mask, depth, pose, etc.
    By leveraging the mask-driven version of the text-to-image SD, which is called the inpainting stable diffusion model, we aim to design a video editing framework given a text prompt and a sequence of mask images corresponding to each frame of the input video. We emphasize that, unlike previous mask-guided inpainting methods \cite{ProPainter, E2FGVI, FGT, DMT}, our approach can deal with masks with arbitrary shapes, positions, and placements in each frame of the input video without any problem. Following the mask-generation strategy presented in \cite{LAMA}, the inpainting diffusion model is resumed from an SD model and fin-tuned for another 200k steps with an additional conditioning signal that provides a mask corresponding to the input image. More specifically, compared to text-to-image stable diffusion models \cite{Imagen, DALLE2, LDMs} which take as input the number of timesteps (diffusion steps) and the text embedding (obtained from a pre-trained text encoder \cite{CLIP}), the existing UNet model in the inpainting stable diffusion model has five additional input channels containing four for the encoded masked image and one for the mask itself. Specifically, these five channels are concatenated with the four ones (timestep and text embedding) used in the text-to-image stable diffusion model. Our approach is to
    first obtain an embedding representation of each frame and its corresponding mask in the input video by using a pre-trained variational autoencoder (VAE). Having the input frame and its mask, we then generate the masked image in which the area that we would like to perform the inpainting task on, has been masked out by a black color. Lastly, by concatenating three signals, we obtain the proper input for the inpainting SD model: 1) Noisy input frame latent: The added noise comes from the last diffusion step of the DDIM inversion stage \cite{DDIM} which is added to the embedding of the frame in the latent space, 2) Masked image latent which is obtained from the encoder of a pre-trained VAE model, and 3) Mask latent which is the down-sampled version of the VAE representation of the mask by using the nearest interpolation of the VAE representation (we suggest the reader to see \textcolor{BlueGreen}{ \cref{fig:Mask_image_processor}} in the supplementary material that visually demonstrates how we achieve a proper input for the inpainting model). After feeding the proper input to the model, in the denoising diffusion stage, the UNet estimates the amount of existing noise in the input. Then, in each denoising step, the model tries to remove the estimated noise from the image latent and obtain a less noisy version of it while editing the masked area in the image by attending to the context of condition signals (mask and text) using existing attention layers in the UNet architecture.
    
    The inpainting SD model gives us more flexibility to control the video content editing by leveraging a user-defined mask (determining a specific region in the input video for the inpainting task) and a text description. Moreover, it allows us to perform different video editing tasks presented in the next section by tuning the guidance scale (GS) parameter \cite{GuidanceScale}. GS guides the model toward any of the two control signals we have: mask and text.
    However, applying the inpainting SD in a naive frame-by-frame manner leads to inconsistencies in frames of the edited frame. Instead, inspired by \cite{TokenFlow, TuneAVideo, Pix2Video}, we systematically redesign the forward path of the pre-trained inpainting diffusion model by replacing the self-attention modules with an extended version of attention modules that induces dependencies between frames. Note that we do not change the architecture of the existing U-Net in the SD model. We manipulate only the computation in the forward path of the self-attention layers. This happens by using several frames instead of one in the computation of self-attention modules to extract similar information or features. That is why our approach does not add any additional training or fine-tuning. Then, having these extracted similar features, we enforce the model to edit (reconstruct) the video in a way that the regions in the frames that have similar features will be kept unchanged while the other parts of the frames will be changed according to the control commands (mask and text prompts). In this way, we ensure that the edited information will be consistent across all the video frames no matter what the shape and position of the masked area is. \cref{fig:DiffusionProcess} (right) demonstrates a visual representation of how the forward path of the extended attention works in our framework. \cref{fig:DiffusionProcess} (left) shows the whole diffusion process of our temporal consistent video editing technique. More specifically, for each diffusion step, similar to \cite{TokenFlow}, we randomly select several frames and their corresponding mask images. Then, these pairs of masks and images are fed into a pre-processor algorithm explained above. Then, the extended attention layers in the U-Net architecture, showing in \cref{fig:DiffusionProcess}, extract similar features from the selected frames. This process is repeated for $T=50$ diffusion steps.


}

\vspace{-5pt}\section{Experiments and Results}\vspace{-5pt}
\label{Experiments_Results}
{
    \begin{figure}
          \centering
          \includegraphics[width=1.02\linewidth]{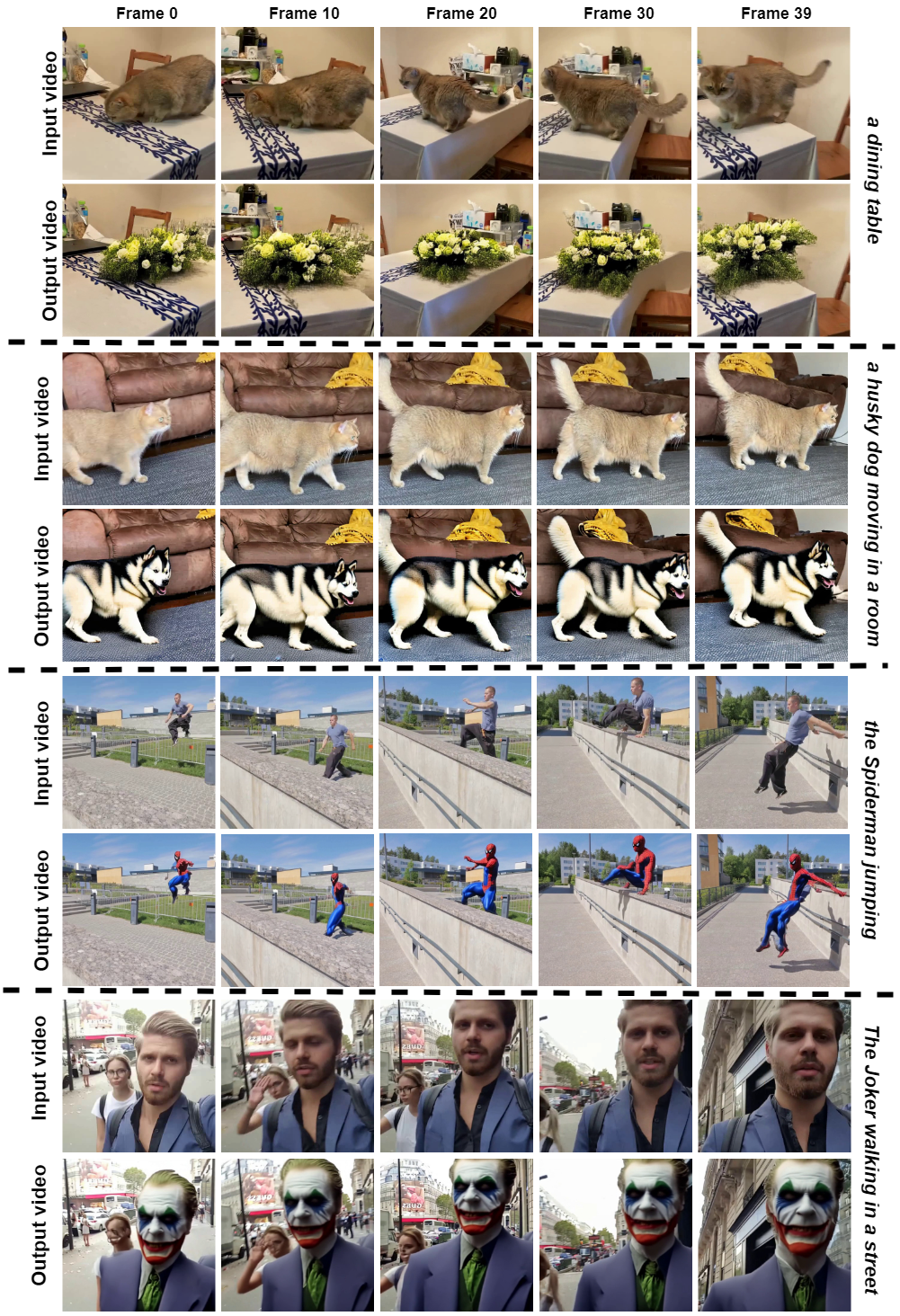}
        
           \caption{A qualitative results of the video object replacement task for different examples\vspace{-15pt}}
           \label{fig:replacement_results_5frame}
    \end{figure}
    
    \begin{figure}
      \centering \includegraphics[width=0.97\linewidth]{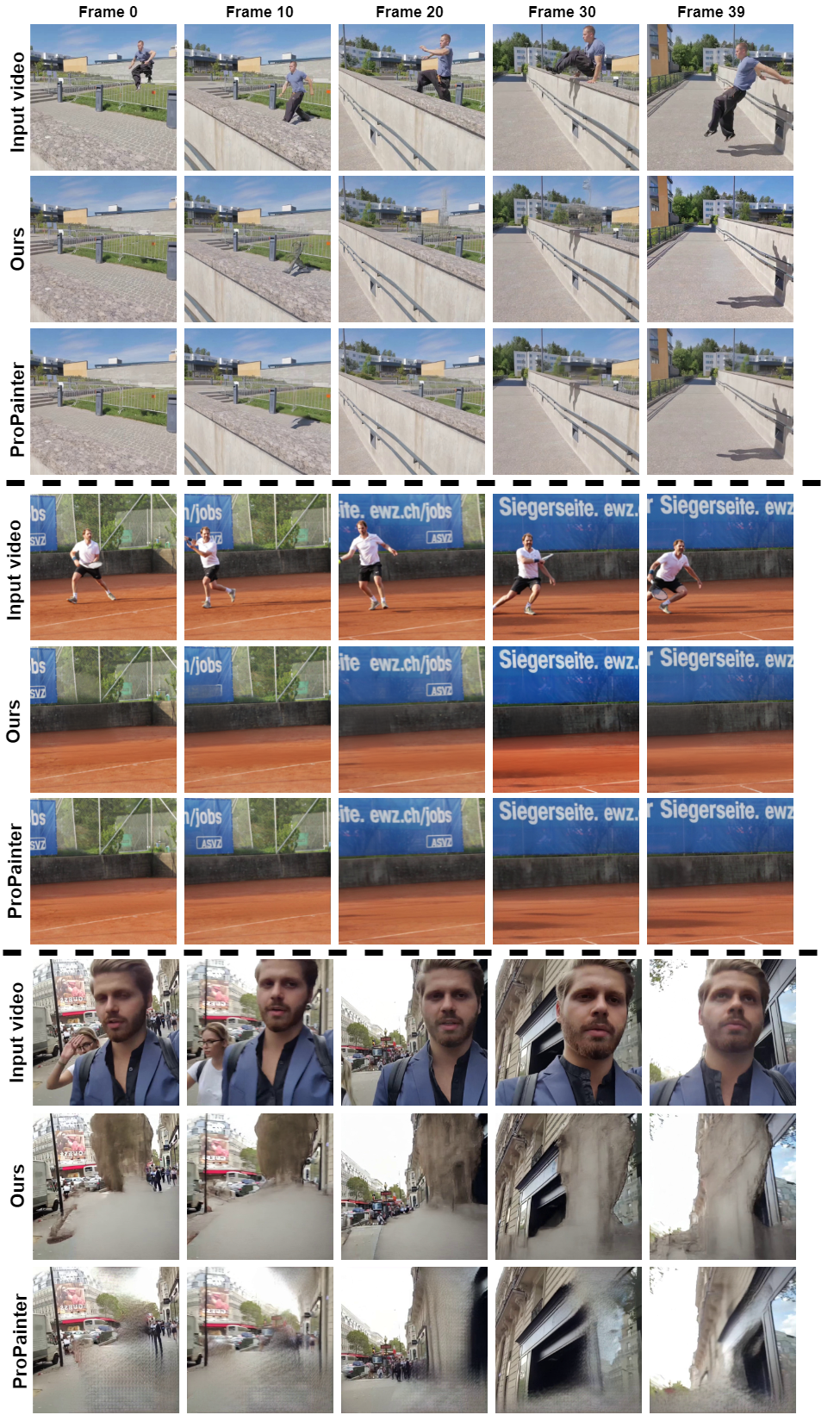}
    
       \caption{A qualitative comparison between ProPainter \cite{ProPainter}, E2FGVI \cite{E2FGVI}, and our proposed method for the object removal task\vspace{-15pt}}
       \label{fig:removal_results_5frame}
    \end{figure}

    \begin{figure}
      \centering
      \includegraphics[width=0.97\linewidth]{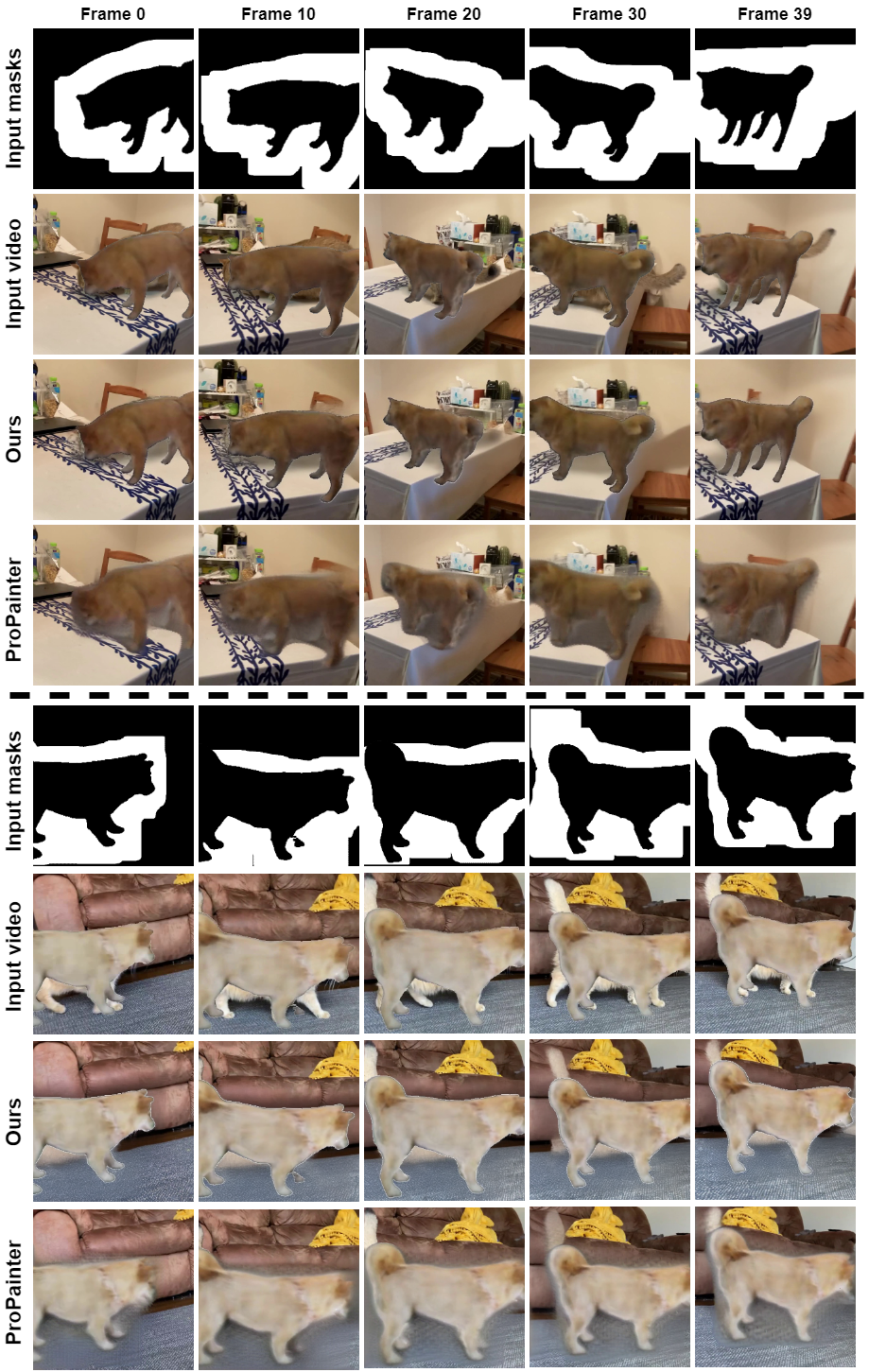}
    
       \caption{A qualitative comparison between ProPainter \cite{ProPainter}, E2FGVI \cite{E2FGVI}, and our proposed method for the object retargeting task \vspace{-17pt}}
       \label{fig:retargeting_results_5frame}
    \end{figure}

    

    To show the effectiveness of our temporal consistent approach, we perform three experiments: consistent retargeting, object replacement, and object removal. We present the comparative results of the proposed method in three subsections in the sequel. More specifically, we qualitatively compare the results of the proposed models, against ProPainter \cite{ProPainter} and E2FGVI \cite{E2FGVI} (in the appendix) for object retargeting and object removal tasks. Also, we show the qualitative results for the object replacement task for different examples. All the experiments are performed on different videos from BANMo \cite{banmo}, DAVIS \cite{DAVIS_Dataset}, and YouTube-VOS \cite{YouTube-VOS_Dataset}. More results on these three tasks are presented in the Appendix.
    
    \noindent \textbf{Video Object Replacement Task}. The goal of this task is to generate an edited video by replacing the foreground with another foreground that adheres to both text and mask prompts while keeping the background the same as the one in the original (input) video. \cref{fig:replacement_results_5frame} showcases the qualitative result of our method on video examples containing different animals and humans with different textual and mask control prompts. As shown, the model can faithfully replace the target object, determined by the mask prompt, with another object which is determined by the text prompt. To the best of our knowledge, our approach is the only one that is able to achieve this task with a mask-based guidance. 

    \noindent \textbf{Video Object Removal Task}. The goal of this task is simply to generate an edited video by replacing the foreground with the background. Almost all mask-guided methods in the literature only try to solve this task. \cref{fig:removal_results_5frame} presents a qualitative comparison of our method against the ProPainter \cite{ProPainter}. For this task, although our method obtains results with the same level of high fidelity compared to state-of-the-art, it still performs acceptably and much better in many other methods in the literature, considering we only fine-tune the parameter guidance scale in our framework without changing any other components or parameters or any fine-tuning or training the model. This enlightens a way to propose a general framework that potentially not only could solve all the discussed tasks together in a unified pipeline, but also obtain state-of-the-art results in this area.

    \noindent \textbf{Consistent Video Object Retargeting}. Object retargeting is a common task in graphics applications \cite{banmo}, that not only removes an object in the scene in a video or an animation sequence, but it replaces it with another object that performs the same action. This comes with significant challenges as now the action needs to be extracted from the data and applied to the new object. However, existing methods extract an object from a video and replace it with an object whose action may no longer be consistent with the video. For example: given a video of two objects (e.g. a dog and a cat) that have been overlaid on top of each other (see the second row of \cref{fig:retargeting_results_5frame}) and a sequence of mask images corresponding to all video frames, the goal of the consistent video object retargeting task is to keep the object which is on top, the \textit{front object}, in the scene and instead remove the other object which lies behind the front object, \textit{behind object} from the scene. More specifically, by processing the masked area, the model tries to remove the \textit{behind object} from the scene in a way that everything in the background becomes temporal consistent across all frames. Our retargetted objects are created using \cite{banmo}. \cref{fig:retargeting_results_5frame} 
    present a comparative analysis of the proposed against state-of-the-art methods including ProPainter \cite{ProPainter} for a video of a dog and cat, focusing on qualitative video editing. In each column, which showcases the information belonging to a specific time in the input video, the first and second rows are the input masks and input frames for that specific time, and other rows demonstrate the result of the edited video when applying our method and ProPainter \cite{ProPainter}, respectively. The results show the superiority of our method to ProPainter \cite{ProPainter} as there are some blurry regions in the results of Propainter \cite{ProPainter} while this is not the case for ours (see more results and also comparison with E2FGVI \cite{E2FGVI} in the appendix). Similar to other mask-guided methods in the literature, the reason ProPainter \cite{ProPainter} and E2FGVI \cite{E2FGVI} fail in this task is that all of them only can handle masks that have a fixed shape, position, and orientation across the frames. However, our method regardless of what the mask is tries to perform the inpainting tasks while achieving the temporal consistency at the same time.
}
\vspace{-7pt}\section{Conclusions}\vspace{-5pt}
\label{Concolusion}
{
    This study introduces a temporal consistent method for video editing with mask and text guidance, relying only on a pre-trained in-painting diffusion model. The proposed approach allows one to use our framework without any additional training or fine-tuning to obtain competitive results on a common object removal task as well as a new object replacement task that prior approaches where existing methods are not viable. 
    As for future work, we believe that the proposed approach can be used for a general set of tasks by enhancing its performance on the object removal task.  
}
{
    \small
    \bibliographystyle{ieeenat_fullname}
    \bibliography{main}
}

\clearpage
\setcounter{page}{1}
\maketitlesupplementary

\section{Quanitative Evaluation}
{
    To numerically evaluate our method, we follow the criteria used by several state-of-the-art approaches in the literature \cite{ProPainter, E2FGVI}: Structural similarity (SSIM) \cite{SSIM} index and peak signal-to-noise ratio (PSNR). These are methods for measuring the similarity between two images. The SSIM index can be viewed as a quality measure of one of the compared images, provided the other image is of perfect quality. We use both PSNR and SSIM metrics to evaluate the visual and perceptual similarity between input and output videos. More specifically, comparing our method with ProPainter \cite{ProPainter} and E2FGvI \cite{E2FGVI} using the metrics mentioned above provides an insight into how similar the corresponding parts are in the input and output (edited) video frames. This is useful when we compare all the methods on the video object removal and video object retargeting task (explained in Section \ref{Experiments_Results} of the main paper). Since in these two tasks it is desired to remove the masked-out areas and replace them with contents around the masked-out area in the input frames. As a result, it is important to know how effective the methods are in editing the masked-out object while keeping other areas in the input video unchanged. Note that we do not compare the results of our method with Propainter and E2FGvI for the video object replacement task as those methods cannot perform those tasks. We achieve better results with all metrics in the video object retargeting task, as indicated in \cref{tab:quantitative_comparison}. The superiority of the proposed method can also be visually verified by the qualitative results in \cref{fig:retargeting_results_5frame} of the main paper. Regarding the results of the video object removal task, our method achieves results more or less the same level of fidelity compared to the state-of-the-art.


    \begin{table}[h]
        \scalebox{0.92}
        {
            \centering
            \begin{tabular}{|c?c|c?c|c|}
                \hline { \textbf{Methods} } & \multicolumn{2}{c?}{ \textbf{Removal Task} } & \multicolumn{2}{c|}{ \textbf{Retargeting Task} } \\
                
                \cline{2-5} & PSNR $\uparrow$ & SSIM $\uparrow$ & PSNR $\uparrow$ & SSIM $\uparrow$ \\
                \hline \textbf{E2FGVI} \cite{E2FGVI} & 18.5454 & \textbf{0.7510} & 19.7115 & 0.7782 \\
                \hline \textbf{ProPainter} \cite{ProPainter}  & 18.1411 & 0.7416 & 20.1209 & 0.7864  \\        
                \hline \textbf{Ours} & \textbf{18.5871} & 0.7467 & \textbf{22.8837}  & \textbf{0.8112} \\
                \hline
            \end{tabular}
        }
        \caption{A quantitative comparison between our proposed method, ProPainter \cite{ProPainter}, and E2FGVI \cite{E2FGVI}}
        \label{tab:quantitative_comparison}
    \end{table}


}

\section{Ablation Studies}
{   
    To show the effectiveness of our design choices, we conduct ablation studies to analyze the underlying mechanisms and components of the proposed methodology by systematically evaluating the effects of removing or modifying specific elements. \cref{fig:ablation_studies} demonstrates ablation experiments with different configurations. Note that we present the results of different ablation experiments on different video examples, and text-mask pair prompts to show the effectiveness and generalizability of our method. We present the ablation studies in four parts in the sequel.

    \begin{figure*}
      \centering
      \includegraphics[width=0.85\linewidth]{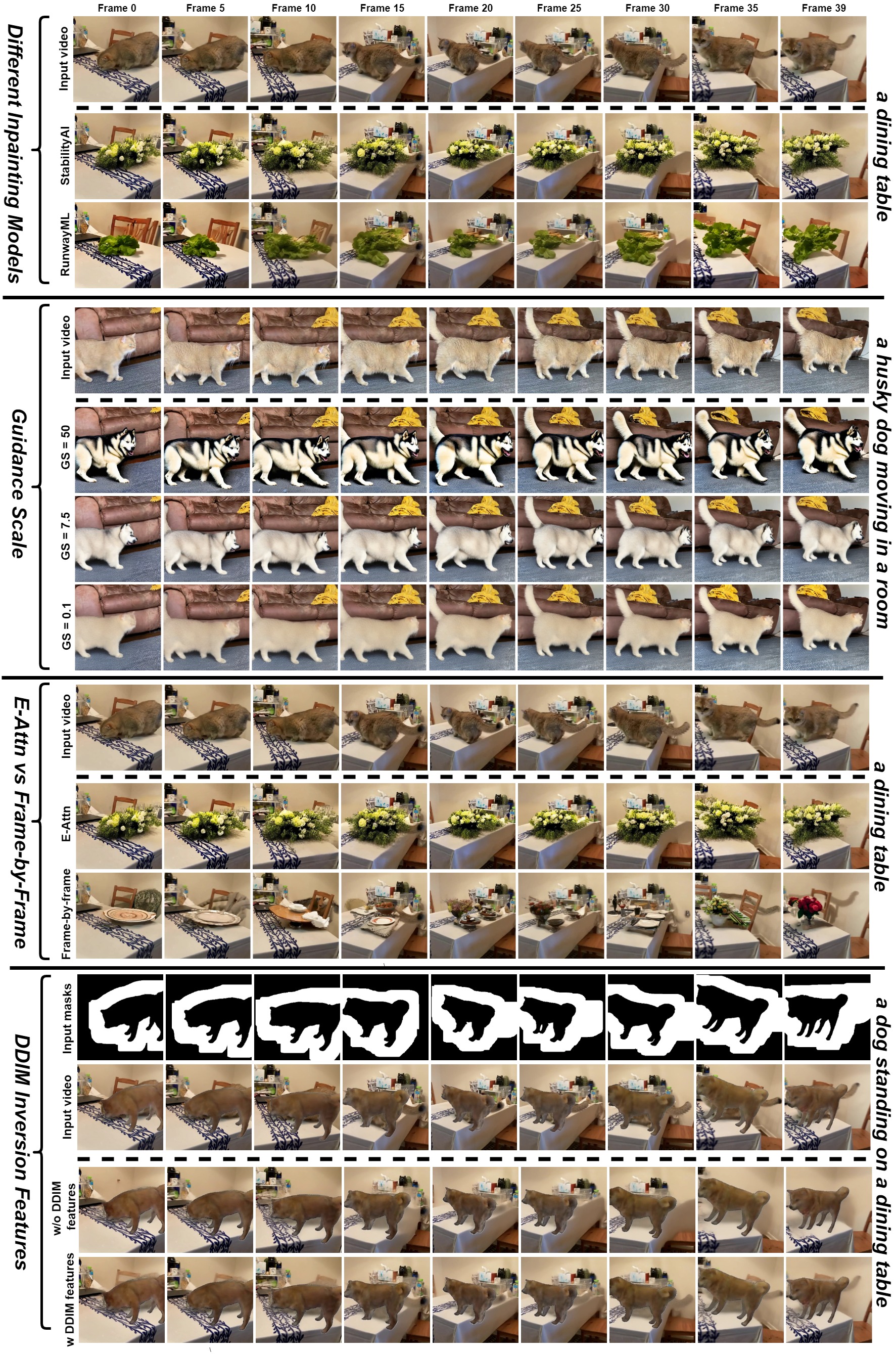}
    
       \caption{Ablation studies on different components and mechanisms of the proposed method.}
           \label{fig:ablation_studies}
    \end{figure*}

    \noindent \textbf{Different inpainting models}. As mentioned in the main paper, we leverage a pre-trained text- and mask-to-image inpainting diffusion model in our methodology. Among all these pre-trained inpainting stable diffusion models, we compare two of the most well-known models developed by StabilityAI \footnote{\href{https://huggingface.co/stabilityai/stable-diffusion-2-inpainting}{Inpainting Stable Diffusion v2 by StabilityAI}} and RunwayML \footnote{\href{https://huggingface.co/runwayml/stable-diffusion-inpainting}{Inpainting Stable Diffusion v1.2 by RunwayML}}. In this experiment, we qualitatively evaluate the performance of these two models in different video environments. In all examples, as shown in \cref{fig:ablation_studies}, the StabilityAI model generates more realistic images with higher quality. Rows \#1-3 in the figure depict several frames of two edited videos resulting from the StabilityAI and RunwayML inpainting diffusion models on the same example with the same mask and text prompts. As shown, the edited video (the flower) is much more realistic than the video produced by the RunwayML model. Note that the depicted frames in \cref{fig:ablation_studies} represent only one example to visualize the quality of these two models in our ablation studies. It is to be noted that our selection of models and examples is based on experimenting with different inpainting models with a sufficient number of examples to compare the quality and sanity of the edited images according to specific text prompts.

    \noindent \textbf{Guidance scale}. We ablate the effect of the {\em guidance scale} (GS) parameter on the video editing qualitative results. Rows \#5-7 of \cref{fig:ablation_studies} show the effect of different GS's on the visual quality of the edited video. According to these results, on the one hand, when the guidance scale is set to a high value, the model strongly adheres to the text prompt. This often leads to images highly aligned with the provided text and mask image prompts, increasing fidelity to the input text (see row \#5 of the figure). On the other hand, a lower guidance scale means the model gives more weight to the noise component during the diffusion process. This results in more diverse and creative outputs, but they are less faithful to the text prompt (see rows \#6, 7 of the figure).

    \noindent \textbf{Extended attention and historical frame information}. To demonstrate the effect of the extended attention module, we qualitatively compare two scenarios for several video examples.
    \begin{itemize}
        \item We replace the self-attention layers in our U-Net architecture with the extended version of the attention layer and use historical frame information to enforce the temporal consistency in the U-Net architecture (we refer to this scenario as \textit{E-Attn} in \cref{fig:ablation_studies}).

        \item We apply the inpainting stable diffusion model in a naive frame-by-frame manner on the input video frames (we call this scenario \textit{frame-by-frame} in \cref{fig:ablation_studies}). As shown in rows \#9 and 10 of the figure, using the extended attention module results in visual consistency in the frames of the edited video while the frames are completely different and inconsistent in the frame-by-frame scenario.
    \end{itemize}

    \noindent \textbf{Features extracted from the DDIM inversion step}. We ablate the effect of incorporating the features extracted from the DDIM inversion step in our temporally consistent video editing procedure. The goal of this step is to obtain a noisy version of each frame by adding a small amount of noise to the less noisy version of the image (the pure image). Existing text-to-video editing methods such as TokenFlow \cite{TokenFlow} utilize these noisy versions of the frames for all diffusion steps obtained from the DDIM inversion process as knowledge to inform the model during the video editing step to keep the content of the edited video as close as possible to the input video. However, for the video inpainting tasks where both mask and text prompts contribute to the edited video, we empirically show that there is no need to incorporate all these noisy versions of the video frames. Instead, the only information required is the noisiest version of video frames (obtained from the last step of the DDIM inversion process) which will serve as an initial value for the denoising step during the editing process of our method. To perform this ablation study, we consider two scenarios.
    
    \begin{itemize}
        \item We remove DDIM inversion features and only consider text-guided, mask-guided, and unguided features during the inference stage.
        \item We concatenate DDIM inversion features with both guided and unguided features.              
    \end{itemize}
    
    The last two rows of \cref{fig:ablation_studies} depict the results of the video object retargeting task in these two scenarios. The results show that removing the DDIM inversion features improves the visual quality of the edited video significantly, while using the DDIM inversion features provides information about the frames of the input video which enforces the method to keep that information in the edited video as well. As a result, some objects/contents (such as the tail of the cat) from the input video can still be observed in the edited frames (the last row of \cref{fig:ablation_studies}) However, this is not a desired effect in applications such as video object removal or retargeting, where it is desired to remove the masked-out area and replace it with other objects in a way that the edited contents are visually consistent and coherent with the non-masked-out area.
}

\section{Modifications to SD Model}
{
    In this section, we outline modifications we made to the text-to-image SD model \cite{LDMs} for temporally consistent object editing in videos.

    \noindent \textbf{Using the inpainting SD model.} The existing text-to-image diffusion models such as SD \cite{LDMs} can only process a textual description as an external command given by a user to manipulate, edit, and synthesize an image. Therefore, to add more control ability to video editing applications, we replace the stable diffusion model recently leveraged by methods such as TokenFlow \cite{TokenFlow} with an inpainting SD model (described in Section \ref{Methodology} of the main paper) which can process both text and mask control commands. Note that this is carried out in the consistent video editing step while we use text-to-image in the DDIM inversion step. The proposed modifications to these two steps are described below.

    \noindent \textbf{Leveraging only the last step of the DDIM inversion process.} While all the noisy frames obtained from the DDIM inversion step (explained in the previous section) are required to generate high-quality videos in recent video editing methods such as TokenFlow \cite{TokenFlow}, we empirically found out that, for video inpainting tasks where both mask and text prompts contribute to the edited video, we need to use the noisy frames generated from only the last step of the DDIM inversion, not all steps (ablation results on this are presented in \cref{fig:ablation_studies}). Intuitively, this is due to the fact that using the mask control command limits the model to edit only the masked-out area in the input video. In this way, the extended attention layers ensure the consistency and coherence of non-masked-out areas by comparing corresponding contents in several random frames of the input video. However, existing methods such as TokenFlow \cite{TokenFlow}, in addition to their consistency methods (no matter what they are), need to incorporate the noisy latent of each input frame to ensure temporal consistency and to keep the information in the non-masked-out areas unchanged as there is no control command other than a textual description. This modification drastically reduces the memory required for our method, as there is no need to compute and process noisy versions of all input video frames in all diffusion steps.    

    \noindent \textbf{Pre-processing input video frames along with text and mask prompts.} Since we replace the text-to-image SD model with the inpainting SD model, we need to provide a meaningful and reliable way to combine and feed the video frames, text, and mask prompts to the inpainting model. To this end, we follow the original implementation of the pre-processor algorithm        \footnote{\href{https://github.com/huggingface/diffusers/blob/main/src/diffusers/pipelines/stable_diffusion/pipeline_stable_diffusion_inpaint.py}{Pipeline stable diffusion inpaint}} \footnote{\href{https://github.com/huggingface/diffusers/blob/main/src/diffusers/image_processor.py}{Image processor}} for the inpainting SD model. We visually demonstrate how we achieve a proper input for the inpainting SD model in \cref{fig:Mask_image_processor}. A detailed explanation of all the pre-processing steps is presented in Section \ref{Methodology} of the main paper.    

    \noindent \textbf{Redesigning the forward path using the extended attention.}
    The core idea and main modification of our method is to systematically redesign the forward path of the pre-trained diffusion model by replacing the self-attention modules with an extended version of attention modules without any additional training or fine-tuning. Applying the extended attention operations on several frames of the input video ensures that the edited information will be consistent across all the video frames. \cref{fig:DiffusionProcess} (right) of the main paper shows exactly how key, query, and value tensors in the extended attention operation are initialized using several randomly selected frames in the input video. This approach allows one to extract similar contents and features across multiple frames of the input video and to keep them unchanged when editing the masked-out area during the inference stage.
    \cref{fig:inference} showcases the functionality of our approach at the inference stage. At a high-level overview, for each diffusion step, similar to \cite{TokenFlow}, we randomly select several frames and their corresponding mask images. Then, these pairs of masks and images are fed into a pre-processor algorithm, demonstrated in \cref{fig:Mask_image_processor}, that processes and combines masks and images. Then, the extended attention layers in the U-Net architecture, shown in \cref{fig:DiffusionProcess} of the main paper, extract similar features from the selected frames. This process is repeated for T = 50 diffusion steps to obtain an acceptable edited video.

    \begin{figure*}
      \centering
      \includegraphics[scale=0.55]{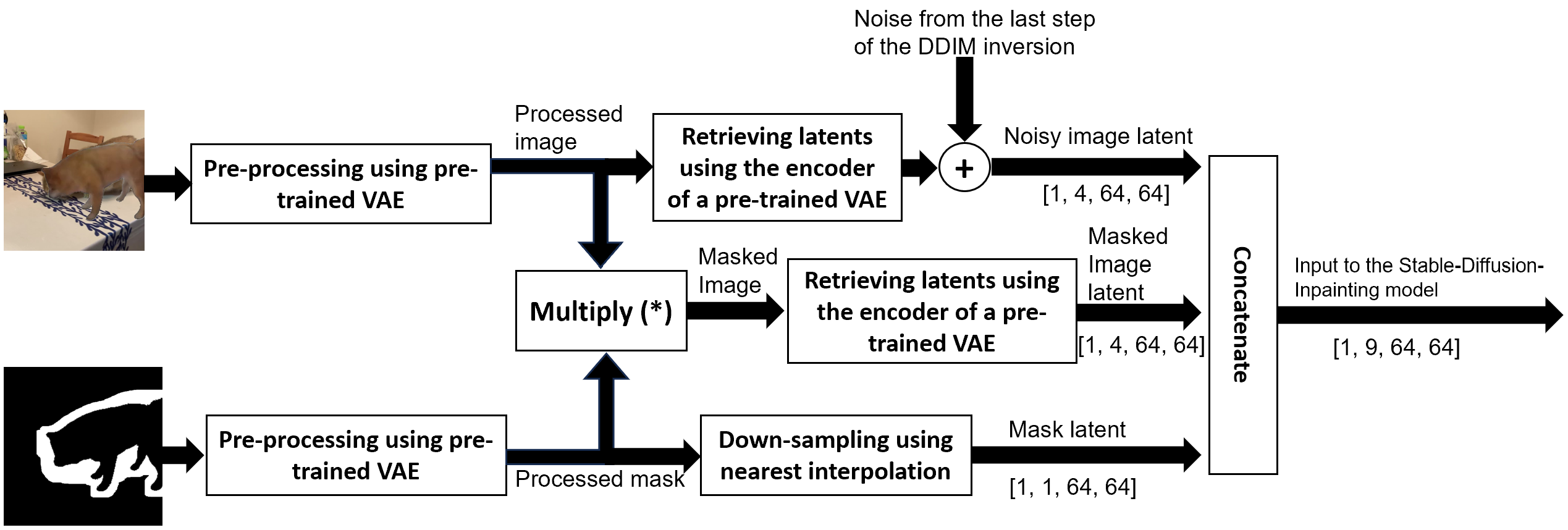}
    
       \caption{The mechanism of pre-processing the input video and its corresponding sequence of masks for the inpainting stable diffusion model}
       \label{fig:Mask_image_processor}
    \end{figure*}
        
    \begin{figure*}
      \centering
      \includegraphics[scale=0.54]{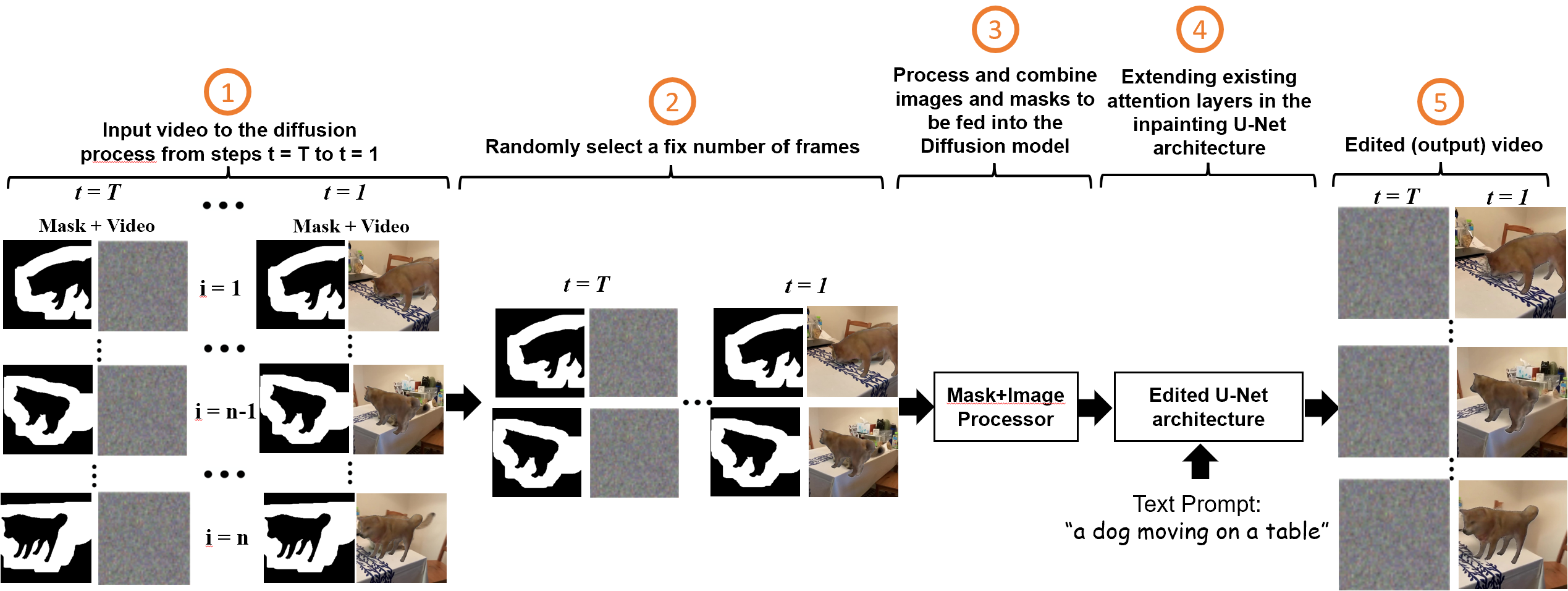}
    
       \caption{An overview of our pipeline at the inference stage}
       \label{fig:inference}
    \end{figure*}

}

\section{Additional Results}
{
    We show expanded results from the main paper featuring more frames and an additional baseline (E2FGVI \cite{E2FGVI}) in \cref{fig:replacement_results_9frame,fig:removal_results_9frame,fig:retargeting_results_9frame}.

        \begin{figure*}
      \centering
      \includegraphics[width=1.01\linewidth]{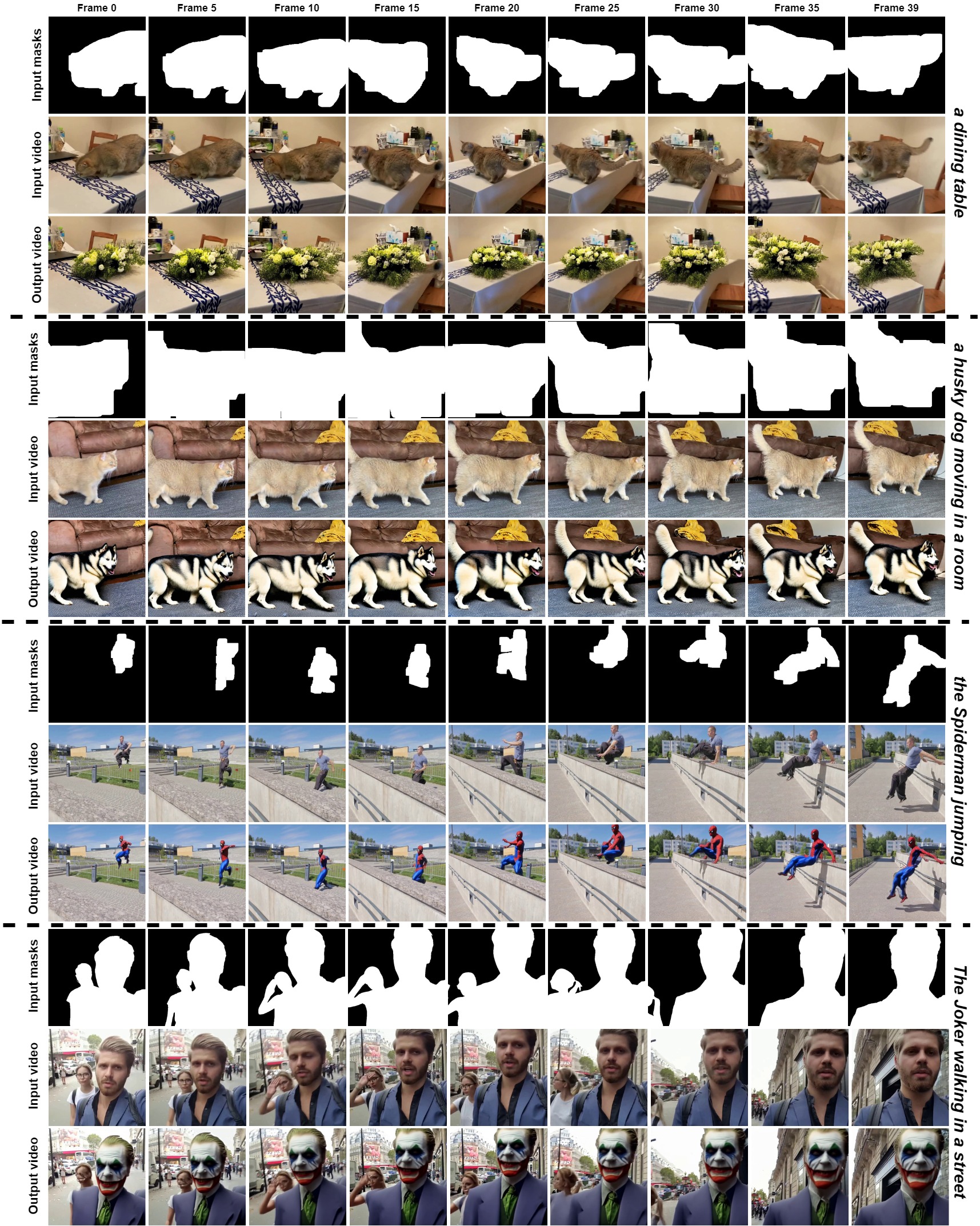}
    
       \caption{A qualitative results of the video object replacement task for different examples}
           \label{fig:replacement_results_9frame}
    \end{figure*}

    \begin{figure*}
      \centering
      \includegraphics[width=1.05\linewidth]{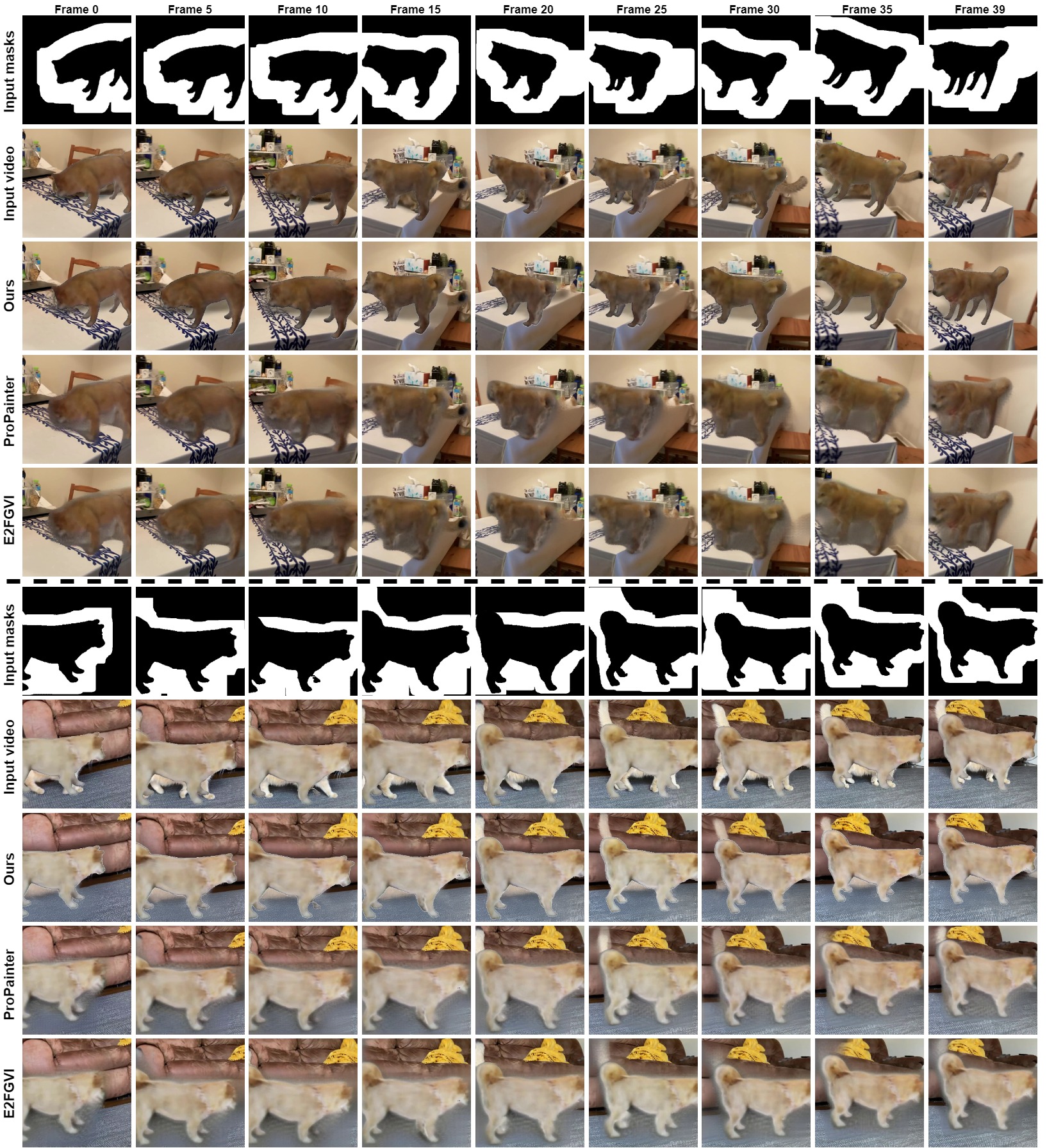}
    
       \caption{A qualitative comparison between ProPainter \cite{ProPainter}, E2FGVI \cite{E2FGVI}, and our proposed method for the object retargeting task}
       \label{fig:retargeting_results_9frame}
    \end{figure*}

    \begin{figure*}
        \centering    \includegraphics[width=0.8\linewidth]{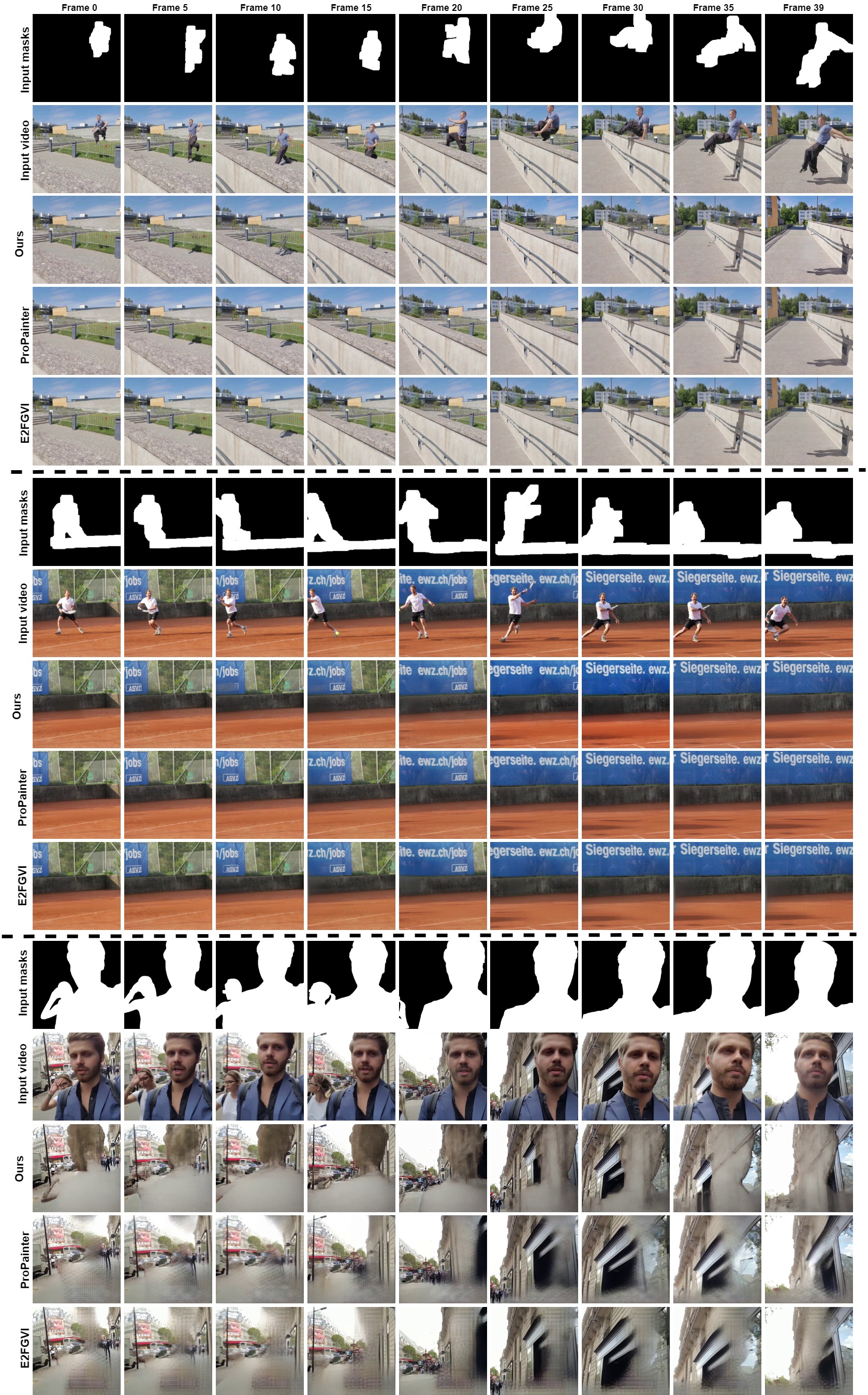}
        
        \caption{A qualitative comparison between ProPainter \cite{ProPainter}, E2FGVI \cite{E2FGVI}, and our proposed method for the object removal task}
        \label{fig:removal_results_9frame}
    \end{figure*}

}

\end{document}